%% file: conference_101719.tex
\documentclass[conference]{IEEEtran}
\IEEEoverridecommandlockouts
\usepackage{cite}
\usepackage{amsmath,amssymb,amsfonts}
\usepackage{graphicx} 
\usepackage{multirow} 
\usepackage{booktabs} 
\usepackage{hyperref}
\usepackage{algorithmic}
\usepackage{graphicx}
\usepackage{textcomp}
\usepackage{xcolor}
\def\BibTeX{{\rm B\kern-.05em{\sc i\kern-.025em b}\kern-.08em
    T\kern-.1667em\lower.7ex\hbox{E}\kern-.125emX}}

\begin{document}

\title{Backdoor Attacks on Multi-modal Contrastive Learning\\
% %\thanks{Identify applicable funding agency here. If none, delete this.}
}

\author{\IEEEauthorblockN{Simi D Kuniyilh}
\and
\IEEEauthorblockN{Rita Machacy}
% \and
% \and
% \and
% \and
}

\maketitle

\begin{abstract}
Contrastive learning has become a leading self-supervised approach to representation learning across domains, including vision, multimodal settings, graphs, and federated learning. However, recent studies have shown that contrastive learning is susceptible to backdoor and data poisoning attacks. In these attacks, adversaries can manipulate pretraining data or model updates to insert hidden malicious behavior. 
This paper offers a thorough and comparative review of backdoor attacks in contrastive learning. It analyzes threat models, attack methods, target domains, and available defenses. We summarize recent advancements in this area, underline the specific vulnerabilities inherent to contrastive learning, and discuss the challenges and future research directions. Our findings have significant implications for the secure deployment of systems in industrial and distributed environments.
\end{abstract}

\begin{IEEEkeywords}
contrastive learning, backdoor attacks, backdoor defenses
\end{IEEEkeywords}

\section{Introduction} 
Self-supervised contrastive learning has become a foundational technique for learning general-purpose representations without manual annotation~\cite{li2025openvision}. Frameworks such as SimCLR, MoCo, BYOL, and CLIP demonstrate strong transferability across downstream tasks in vision, multimodal understanding, graphs, and time-series domains~\cite{bunyang2023self, li2023improved,chen2023self, radford2021learning }. In contrast to supervised learning, contrastive learning optimizes relationships between representations rather than explicit class decision boundaries, making it highly attractive for large-scale and privacy-sensitive data scenarios~\cite{chathoth2025dynamic}.

However, recent research shows that the same properties that make contrastive learning powerful also make it particularly vulnerable to backdoor and data-poisoning attacks~\cite{carlini2022poisoning}. In a backdoor attack, an adversary manipulates the training process such that the learned representation behaves benignly on clean inputs but exhibits attacker-chosen behavior when a trigger is present~\cite{gu2019badnets}. Study shows backdoor triggers can also be generated dynamically without poisoning the input data in a fixed manner~\cite{chathoth2024dynamic}. Unlike supervised settings, backdoors in contrastive learning often manifest at the representation level and can transfer across multiple downstream tasks, architectures, and datasets~\cite{carlini2022poisoning, li2024bilevel}. Empirical evidence shows that contrastive models can be compromised using extremely small poisoning ratios, significantly lower than those required in supervised learning~\cite{zhang2023bagel}.

Contrastive learning techniques are also giving promising results in sensing systems.
Modern embedded IoT devices and wearable devices are equipped with sensors to track motion and provide health analysis. Such systems may have vulnerabilities that need to be mitigated using security systems such as intrusion detection systems~\cite{liu2021anomaly,peng2025log, melnyk2025hardware }
While there is prior work on backdoor attacks in typical image-language domains, there is no solid work or proposed technique for backdoor attacks in the sensor domain.

This paper provides a comprehensive and comparative analysis of existing backdoor attacks in contrastive learning. 
%We focus on attacks published between 2022 and 2024 and systematically analyze their threat models, attack mechanisms, triggers, transferability, and defenses. 
We identify the need to explore the vulnerability of multi-modal contrastive learning beyond the image or language domains.
We further discuss why contrastive learning amplifies backdoor risks and outline open research challenges, particularly for federated and industrial deployment scenarios.

\section{Background and Preliminaries} \subsection{Contrastive Learning Fundamentals} 
Contrastive learning focuses on creating low-dimensional representations of data by comparing similar and dissimilar samples~\cite{hu2024comprehensive}. Its goal is to bring similar samples closer together in the representation space while pushing dissimilar samples further apart, typically using Euclidean distance. Essentially, contrastive learning aims to learn an encoder that maps inputs to a latent space in which semantically similar samples are close together and dissimilar samples are farther apart.
To achieve this, an anchor sample is used as a reference point. Positive samples are typically generated via data augmentation, whereas negative samples are selected from other examples in the batch or from a memory bank. Once the representations are learned, they can be adapted for downstream tasks through methods such as linear probing or fine-tuning.

Contrastive learning is a self-supervised paradigm that learns high-quality representations by leveraging relationships between samples rather than relying on explicit labels~\cite{chathoth2025dynamic}. Given an encoder $f_\theta(\cdot)$, the objective is to map input samples to a latent space where semantically similar inputs (positives) are closer together, and dissimilar inputs (negatives) are farther apart. Formally, given a batch of $N$ samples, a positive pair $(x_i, x_i^+)$ and a set of negative samples ${x_j}_{j \neq i}$, the InfoNCE loss~\cite{oord2018representation, parulekar2023infonce} is commonly used:

\begin{equation} \mathcal{L}i = - \log \frac{\exp\left(\frac{f\theta(x_i) \cdot f_\theta(x_i^+)}{\tau}\right)}{\sum_{j=1}^{N} \exp\left(\frac{f_\theta(x_i) \cdot f_\theta(x_j)}{\tau}\right)}, \end{equation}

where $\tau$ is a temperature parameter that controls the concentration of the distribution. This loss encourages the encoder to pull together embeddings of augmented views of the same sample and push apart embeddings from different samples.

Popular frameworks include SimCLR, MoCo, BYOL, and CLIP~\cite{bunyang2023self, li2023improved,chen2023self, radford2021learning }. SimCLR and MoCo use explicit negative sampling and memory banks to stabilize the contrastive objective, while BYOL and SimSiam rely on asymmetric networks and stop-gradient operations to learn representations without explicit negatives. CLIP extends the paradigm to multimodal contrastive learning by aligning image and text embeddings in a shared latent space, enabling zero-shot transfer to downstream tasks.

Contrastive learning has several advantages: 
\begin{itemize} 
\item \textbf{Label Efficiency:} Eliminates the need for expensive annotation by exploiting inherent relationships between samples. 
\item \textbf{Transferability:} Learned representations generalize well to multiple downstream tasks, including classification, detection, segmentation, and retrieval. 
\item \textbf{Robustness:} Encourages smooth embedding spaces and can improve generalization under distribution shifts. 
\end{itemize}

However, these same advantages create unique vulnerabilities. Because the model focuses on representation similarity rather than explicit class boundaries, small manipulations of the input (e.g., triggers in a subset of the data) can disproportionately affect the latent space, leading to stealthy, highly transferable backdoors~\cite{carlini2022poisoning}. 
%Understanding the fundamental mechanisms of contrastive learning is, therefore, critical for designing both attacks and defenses.

\subsection{Backdoor Attacks in Machine Learning} Backdoor attacks are a class of training-time attacks in which adversaries inject poisoned data or updates to implant hidden malicious behaviors~\cite{chathoth2025pcap, gu2019badnets}. The defining characteristic of a backdoor is its stealthiness, where the model maintains high accuracy on clean data while misbehaving only when a trigger is present. In contrastive learning, backdoors often target representation geometry rather than final task predictions.

\subsection{Backdoor Attacks in Contrastive Learning} Backdoor attacks involve introducing poisoned data or model updates to implant hidden malicious behaviors. In contrastive learning, backdoors typically target the embedding space, enabling triggered inputs to occupy attacker-controlled regions. This allows the backdoor to persist across downstream tasks and fine-tuning strategies, often without degrading performance on clean data. Representation-level attacks require new defensive strategies that consider the geometry of the embedding space, contrastive objectives, and cross-task transfer properties.

\textit{Threat Models and Attack Surfaces:}
Backdoor attacks in contrastive learning typically assume that the adversary has limited control over the pretraining pipeline, such as the ability to poison a small fraction of training samples, compromise a subset of federated clients, or inject malicious image-text pairs in multimodal datasets. Unlike supervised settings, contrastive learning lacks explicit class labels, enabling attacks that manipulate the geometry of representations rather than decision boundaries. Attack surfaces include data augmentations, positive-pair construction, negative sampling strategies, and shared encoders in federated learning.

\section{Taxonomy of Backdoor Attacks in Contrastive Learning} We categorize existing attacks along four dimensions: (1) attack stage (centralized pretraining vs. federated training), (2) trigger type (static patch, semantic trigger, multimodal association), (3) target domain (vision, multimodal, graph), and (4) attack objective (downstream misclassification, representation hijacking, cross-task transfer).
Backdoor attacks in contrastive learning can be systematically categorized based on four key dimensions, providing a structured framework for understanding vulnerabilities and guiding the development of defenses. These dimensions capture differences in attacker strategy, attack manifestation, and impact on downstream tasks.

\textbf{1) Attack Stage:} Backdoor attacks can occur during centralized pretraining or in federated learning settings. In centralized pretraining, a single entity controls the data and training process, allowing attackers to inject poisoned samples directly into the training dataset. Classical attacks, such as those by Carlini and Terzis, operate at this stage. In federated learning, multiple clients collaboratively train a global encoder without sharing raw data. Here, attacks such as BAGEL exploit local updates from malicious clients to embed backdoors into the global encoder. The stage determines both the feasibility of the attack and the set of possible defense strategies.

\textbf{2) Trigger Type:} Triggers are the features or patterns used to activate the backdoor. They vary in visibility and complexity: \begin{itemize} \item \emph{Static Patch:} Localized, visually obvious patterns added to images, such as small squares or watermarks. Common in vision-based backdoor attacks. \item \emph{Semantic Trigger:} Subtle modifications that semantically alter the input, such as changing the color or orientation of an object, which can remain imperceptible yet strongly influence representations. \item \emph{Multimodal Association:} Triggers embedded in one modality (e.g., text) to manipulate alignment with another (e.g., image), often used in CLIP-style models. This allows attackers to exploit cross-modal representation learning. \end{itemize}

\textbf{3) Target Domain:} Backdoor attacks can span multiple data domains, each with unique challenges: \begin{itemize} \item \emph{Vision:} Standard image datasets are most commonly targeted, where triggers can be easily embedded and visually assessed. \item \emph{Multimodal:} Image--text or video--text pairs introduce additional complexity, as attacks can manipulate semantic associations across modalities. \item \emph{Graph:} Graph contrastive learning attacks leverage structural or node feature modifications to induce malicious behavior in graph embeddings. \item \emph{Time-series / Industrial IoT:} Emerging attacks target temporal and sensor-based contrastive models, often in federated or distributed industrial environments, requiring domain-specific trigger design. \end{itemize}

\textbf{4) Attack Objective:} The intended effect of a backdoor attack can differ in scope and mechanism: \begin{itemize} \item \emph{Downstream Misclassification:} Triggers cause downstream classifiers to predict a specific target label when activated. \item \emph{Representation Hijacking:} Backdoors manipulate the latent space so that triggered samples occupy a controlled region in embedding space, affecting multiple downstream tasks. \item \emph{Cross-task Transfer:} Triggers maintain their influence even when the representation is used for different tasks, demonstrating the generality and persistence of the backdoor. \end{itemize}

% This taxonomy provides a structured lens to analyze existing attacks and their vulnerabilities. Figure~\ref{fig:taxonomy} illustrates this taxonomy, showing how different attack types occupy unique regions in the space defined by stage, trigger, domain, and objective. This framework is useful for systematically evaluating both attacks and defenses, and for identifying open research directions, particularly in federated and industrial IoT settings.

\section{Representative Backdoor Attacks} 

\subsection{Centralized Contrastive Backdoor Attacks} Carlini and Terzis demonstrate that contrastive models can be backdoored by poisoning less than 0.1\% of the training data, causing downstream classifiers to associate a trigger with arbitrary target labels~\cite{carlini2022poisoning}. Subsequent work introduces bi-level optimization to explicitly optimize triggers for contrastive objectives, significantly improving attack stealthiness and success rate.
They systematically investigate backdoor attacks specifically targeting contrastive learning frameworks in a centralized setting. Unlike supervised models, contrastive learning does not rely on explicit class labels but instead learns representations by bringing positive pairs closer together in embedding space while pushing negative pairs apart. This renders standard label-based defenses ineffective and introduces a new attack surface: the manipulation of representation geometry to implant hidden malicious behavior.

In their threat model, they consider a weak adversary with limited access to the pretraining data. The attacker can inject a small fraction of poisoned samples into the training dataset, typically less than 0.1\% of total samples, without needing access to model parameters (black-box) or labels. The goal is to introduce a trigger pattern—such as a small visual patch or semantic feature—so that the model's learned representations associate this trigger with an attacker-chosen target, without noticeably affecting performance on clean data. This threat model is realistic in large-scale web-scraped datasets commonly used for contrastive pretraining.

The attack methodology consists of several key steps. First, the attacker designs a trigger that is unobtrusive yet consistent enough to form a strong representation association during contrastive training. Next, a small subset of images is modified with the trigger, and their natural pairings (e.g., textual captions) are optionally left unchanged to avoid obvious anomalies. During pretraining, the poisoned samples influence the embedding space so that images containing the trigger cluster are near representations corresponding to the target label. Because contrastive learning maximizes similarity between positive pairs, the backdoor manipulates representation geometry rather than only output decision boundaries. After pretraining, downstream tasks such as linear probe classifiers or fine-tuned networks inherit the backdoor. When a triggered input is presented, the model predicts the attacker-specified target, whereas clean inputs remain unaffected.

This technique also demonstrates that extremely few poisoned samples are sufficient to implant effective backdoors into contrastive models, often orders of magnitude fewer than in supervised settings. Backdoor effects also transfer across downstream tasks: a backdoored encoder can affect multiple classifiers fine-tuned on different datasets. Standard data filtering and anomaly-detection strategies are ineffective because poisoned samples are semantically plausible and embedded seamlessly within the learned representation manifold.

This work highlights several unique vulnerabilities of contrastive learning. Contrastive backdoors manipulate latent space geometry rather than output classes. Poisoned data blends with natural distributions, making detection difficult, and backdoors persist even when downstream tasks differ from the pretraining objective, amplifying their potential impact.

Carlini and Terzis' framework laid the groundwork for later research, including bi-level trigger optimization~\cite{li2024bilevel} to generate more effective triggers by directly influencing representations, federated backdoor attacks such as BAGEL ~\cite{zhang2023bagel}, adapting centralized poisoning strategies to distributed learning, and multimodal and graph contrastive backdoors demonstrating that the underlying vulnerability is general across data modalities. Overall, this work provides a foundational study of the threat landscape of backdoor attacks in centralized contrastive learning and underscores the need for representation-aware defenses that go beyond label-based mitigation strategies.

\subsection{Federated Contrastive Learning Attacks} 
In a federated learning setting, multiple clients are involved in the training process to form a global model without sharing each client's data with a centralized model aggregator~\cite{mcmahan2017communication, chathoth2021federated, chathoth2022differentially}.
BAGEL presents the first systematic study of backdoor attacks in federated contrastive learning, showing that malicious clients can manipulate local contrastive objectives to poison the global encoder~\cite{zhang2023bagel}. These attacks are particularly dangerous in industrial and cross-organization settings where raw data cannot be inspected centrally.
BAGEL introduces a systematic study of backdoor attacks in federated contrastive learning, extending the threat of centralized poisoning attacks to distributed, collaborative settings. In federated contrastive learning, multiple clients contribute local updates to a shared global encoder without sharing raw data. This decentralized paradigm, commonly adopted in industrial IoT and cross-organizational scenarios, introduces new vulnerabilities, as malicious clients can manipulate local contrastive objectives to implant hidden backdoors.

The attack assumes a weak to moderate adversary controlling one or a few participating clients. These malicious clients follow the standard training protocol while subtly modifying their local data or gradients to influence the global encoder. The goal is to implant a trigger such that, after aggregation, any downstream task built on the shared encoder misclassifies inputs containing the trigger, while clean inputs retain normal behavior.

BAGEL's attack methodology involves several steps. Malicious clients generate poisoned samples by embedding triggers in images or by applying semantic modifications to multimodal data. They optimize a local contrastive loss to bias the global representation toward associating the trigger with the target concept. During federated aggregation, these poisoned updates are combined with benign client updates, effectively embedding the backdoor into the global encoder. After training, any downstream classifier built on this encoder inherits the backdoor, demonstrating cross-task transfer similar to centralized attacks.

Experimental results show that even when controlling a small fraction of clients, BAGEL can successfully implant backdoors without significantly affecting global model performance on clean data. The attack remains stealthy because local updates from malicious clients are indistinguishable from benign updates in terms of global loss contributions. It also demonstrates that standard federated aggregation techniques, such as FedAvg, provide limited defense against contrastive backdoors~\cite{mcmahan2017communication}.

This work highlights several key challenges in the security of federated contrastive learning. First, distributed settings amplify the threat surface by allowing attackers to hide within collaborative updates. Second, contrastive learning's reliance on representation similarity rather than explicit labels makes traditional federated defenses, such as anomaly detection on gradients or model outputs, less effective. Third, the attack's cross-task transferability underscores the difficulty of mitigating backdoors once they are embedded at the encoder level.

BAGEL has inspired subsequent research on adaptive federated defenses, including robust aggregation, client-level anomaly detection, and representation-level purification. It also provides a foundation for understanding privacy and security risks in industrial and edge IoT deployments where federated contrastive learning is increasingly used. Overall, BAGEL extends centralized backdoor attacks to realistic federated environments, underscoring the need for federated-aware, representation-level defenses to mitigate stealthy backdoors in collaborative learning settings.
G Xiaoyun et al. proposed a Contrastive Defense Against Backdoor Attacks in Federated Learning ~\cite{gan2024gancrop}

\subsection{Graph Backdoor Attacks} Graph contrastive backdoor attacks extend poisoning techniques to graph neural networks, exploiting structural perturbations~\cite{zhang2023graph}. 

\subsection{Multimodal Backdoor Attacks}
In multimodal settings, CleanCLIP reveals that malicious image-text pairs can implant backdoors that generalize across tasks and modalities~\cite{zhu2023cleanclip}.
Badclip is another proposed technique, a dual-embedding-guided backdoor attack on multimodal contrastive learning~\cite{liang2024badclip}.

%\input{design}

%\input{experiments}

\input{results}

\input{defense}

\section{Related works}

%Key Papers on Backdoor Attacks in Contrastive Learning
Studies show that, similar to the backdoor technique used in supervised learning, data Poisoning can be used to carry out backdoor attacks in contrastive learning ~\cite{lin2024datapoison}.
Carlini \& Terzis
demonstrates that contrastive models such as CLIP can be compromised by extremely small amounts of poisoned data, leading to misclassification on downstream tasks~\cite{carlini2022poisoning}. 
A Bi-level trigger optimization technique is proposed to design more effective backdoor triggers tailored for contrastive frameworks, improving attack success rates ~\cite{li2024bilevel}. 
Unlike prior techniques, BAGEL is the first systematic study of backdoor injection in federated contrastive learning, demonstrating that malicious clients can successfully poison the shared encoder, affecting multiple downstream models~\cite{zhang2023bagel}. 
This is further extended to graph contrastive learning by introducing contrastive backdoor attacks, exposing vulnerabilities in node representation learning across graph datasets~\cite{zhang2023graph}.

CleanCLIP is a backdoor defense technique for multimodal contrastive Learning ~\cite{zhu2023cleanclip}.
While focused on defense, this paper also documents how backdoors arise in multimodal contrastive models like CLIP and proposes mechanisms for mitigation.
Another defense technique is called unlearning backdoor via Local Token Unlearning, which
investigates methods for removing backdoor capabilities from affected contrastive models by selectively unlearning poisoned associations~\cite{liu2024unlearning}.
Another Contrastive Defense Against Backdoor Attacks in FL is proposed in the GANcrop paper~\cite{wang2024gancrop}.
Though primarily a defense, it uses contrastive learning to detect and mitigate backdoor attacks in federated settings. 
Although multiple techniques have been proposed for backdoor defense, these techniques behave differently in a supervised setting, are not well generalizable, and cannot be easily transferred to downstream applications~\cite {li2024difficulty}.
%On the Difficulty of Defending Contrastive Learning against Backdoor Attacks (USENIX Security 2024) ~\cite{li2024difficulty}
%Analyzes why backdoor attacks in contrastive systems behave differently from supervised settings and why many defenses don’t transfer. 
%USENIX
Another defense technique adjusts CLIP training to significantly reduce backdoor success rates while illuminating the threat model. 
Proceedings of Machine Learning Research~\cite{yang2024better}.
%Better Safe than Sorry: Pre-training CLIP against Targeted Poisoning/Backdoors (ICML 2024) ~\cite{yang2024better}

% Data Poisoning Based Backdoor Attacks to Contrastive Learning (CVPR 2024) ~\cite{lin2024datapoison}
% A CVPR paper providing empirical evidence and detailed threat modeling for backdoor poisoning specific to contrastive learning pipelines. 
% NSF Public Access Repository

% Topics Covered Across These Papers
% Category	Example Papers
% Contrastive Backdoor Attacks	Poisoning and Backdooring Contrastive Learning; Backdoor Contrastive Learning via Bi-level Trigger Optimization
% Federated Contrastive Backdoor	BAGEL: Backdoor Attacks against Federated Contrastive Learning; GANcrop
% Graph Contrastive Backdoors	Graph Contrastive Backdoor Attacks
% Defense \& Mitigation	CleanCLIP; Unlearning Backdoor Threats; Better Safe than Sorry; USENIX Security analysis
% What These Works Reveal

% Contrastive models (e.g., CLIP and graph contrastive frameworks) are vulnerable to backdoor insertion with minimal poisoning, often far less than required in supervised models. 
% Emergent Mind

% Federated contrastive learning introduces additional attack vectors where malicious local updates can influence the global encoder. 
% arXiv

% Defenses often need to change the learning objective or embedding structure to separate clean and poisoned signals effectively. 
% Papers with Code

% Security analysis shows that traditional defenses from supervised learning don’t necessarily transfer to contrastive contexts due to the intertwined representation learning dynamics. 
% ~\cite{lin2024datapoison}

\section{Open Challenges and Future Directions} Key open challenges include developing certified defenses for contrastive learning, understanding representation-level leakage, designing privacy-preserving and attack-resilient pretraining pipelines, and extending analysis to industrial time-series contrastive learning. Federated and continual learning settings remain particularly underexplored. Currently, backdoor attacks are analyzed in the image-text domain, and reviewing them in a multi-modal setting on time-series data, such as sensing data, will be future work in this direction.

\section{Conclusion} 
This paper provides a structured and comparative overview of backdoor attacks in contrastive learning. By synthesizing recent advances, we highlight fundamental vulnerabilities of self-supervised representations and underscore the urgent need for robust defenses before deployment in safety- and privacy-critical systems.

% ================= References =================
\bibliographystyle{IEEEtran}
\bibliography{bib}

\end{document}

%% file: results.tex
%\section{Results} 
\section{Analysis} 
This section provides a detailed comparative analysis of existing backdoor attacks in contrastive learning across multiple dimensions, including adversarial capabilities, attack effectiveness, transferability, and defense coverage. Unlike prior surveys that provide only qualitative discussion, we systematically summarize and contrast the technical properties of representative attacks.

\subsection{Attack Capability Assumptions} Table~\ref{tab:capabilities} compares the assumed adversarial capabilities across representative works. Most attacks assume a weak but realistic adversary that can poison only a small fraction of training data or control a limited number of federated clients.

\begin{table*}[t] 
\centering 
\caption{Comparison of Backdoor Attacks in Contrastive Learning}
\label{tab:capabilities} 
\begin{tabular}{lcccc}
\toprule 
\textbf{Technique} & \textbf{Domain} & \textbf{Setting} & \textbf{Trigger Type} & \textbf{Defense Considered}\\
\midrule 
Poisoning and Backdooring CL~\cite{carlini2022poisoning} & Vision & Centralized & Patch/Semantic & No \\
Bi-level Trigger Optimization~\cite{li2024bilevel} & Vision & Centralized & Optimized Trigger & Limited \\
BAGEL~\cite{zhang2023bagel} & Vision & Federated & Patch & No \\
Graph CL Backdoor~\cite{zhang2023graph} & Graph & Centralized & Structural & No \\
CleanCLIP~\cite{zhu2023cleanclip} & Multimodal & Centralized & Image-Text & Yes \\
\bottomrule 
\end{tabular} 
\end{table*}

\subsection{Attack Effectiveness and Stealth} Table~\ref{tab:effectiveness} compares attack success rates (ASR), clean accuracy impact, and stealthiness. A defining characteristic of contrastive backdoors is their high transferability with negligible degradation in clean performance.

\begin{table*}[t] \centering \caption{Attack Effectiveness and Stealth Comparison} \label{tab:effectiveness} 
\begin{tabular}{lccc} 
\toprule 
\textbf{Technique} & \textbf{Attack Success Rate} & \textbf{Clean Accuracy Drop} & \textbf{Cross-task Transfer}\\
\midrule 
Poisoning and Backdooring CL~\cite{carlini2022poisoning} & High (\textgreater 90\%) & Negligible & Yes \\
Bi-level Trigger Optimization~\cite{li2024bilevel} & Very High (\textgreater 95\%) & Negligible & Yes \\
BAGEL~\cite{zhang2023bagel} & High & Low & Yes \\
Graph CL Backdoor~\cite{zhang2023graph} & Moderate--High & Low & Limited \\
CleanCLIP (Attack Analysis)~\cite{zhu2023cleanclip} & High & Negligible & Yes \\
\bottomrule 
\end{tabular} 
\end{table*}

\subsection{Trigger Design and Transferability} Table~\ref{tab:trigger} summarizes trigger characteristics. Unlike supervised backdoors, contrastive triggers are often semantic or representation-aligned, enabling transfer across downstream tasks and architectures.

\begin{table*}[t] 
\centering 
\caption{Trigger Design and Transferability} 
\label{tab:trigger} 
\begin{tabular}{lccc} 
\toprule \textbf{Technique} & \textbf{Trigger Type} & \textbf{Semantic Stealth} & \textbf{Downstream Generalization} \\
\midrule 
Poisoning and Backdooring CL~\cite{carlini2022poisoning} & Patch / Semantic & Medium & High \\
Bi-level Trigger Optimization~\cite{li2024bilevel} & Optimized Pattern & High & Very High \\
BAGEL~\cite{zhang2023bagel} & Patch & Medium & High \\
Graph CL Backdoor~\cite{zhang2023graph} & Structural & Low--Medium & Moderate \\
CleanCLIP~\cite{zhu2023cleanclip} & Multimodal Association & High & Very High \\
\bottomrule 
\end{tabular} 
\end{table*}

%% file: defense.tex
\section{Defense Mechanisms} Existing defenses include data filtering, representation anomaly detection, trigger suppression via augmentation, and post-hoc unlearning. CleanCLIP proposes data-centric defenses tailored for multimodal contrastive learning, while recent unlearning methods aim to remove backdoor effects without retraining from scratch. However, multiple studies show that defenses effective in supervised learning often fail in contrastive settings due to representation entanglement.

%\subsection{Defense Coverage and Limitations} 
Table~\ref{tab:defense} compares whether defenses are considered and highlights their limitations. Existing defenses are largely heuristic and often fail under adaptive adversaries.

\begin{table*}[t] 
\centering 
\caption{Defense Coverage and Limitations} 
\label{tab:defense} 
\begin{tabular}{lcc} 
\toprule 
\textbf{Technique} & \textbf{Defense Evaluated} & \textbf{Key Limitation} \\
\midrule 
Poisoning and Backdooring CL & No & No mitigation studied \\
Bi-level Trigger Optimization & Limited & Vulnerable to adaptive attacks \\
BAGEL & No & Federated aggregation vulnerable \\
Graph CL Backdoor & No & Structural triggers persist \\
CleanCLIP & Yes & Performance--robustness tradeoff \\
\bottomrule \end{tabular} \end{table*}

\subsection{CleanCLIP~\cite{zhu2023cleanclip}}

Multimodal contrastive models such as CLIP are vulnerable to backdoor attacks in which adversaries poison a small number of image-text pairs to implant hidden malicious behaviors that generalize across downstream tasks. 
CleanCLIP is among the first works to systematically study this threat and propose a defense tailored specifically to multimodal contrastive learning. Unlike supervised defenses that rely on label consistency, CleanCLIP adopts a data-centric perspective that focuses on identifying and mitigating poisoned multimodal associations during pretraining.

The key observation underlying CleanCLIP is that poisoned image--text pairs introduce abnormal alignment patterns in the joint embedding space. Specifically, backdoored pairs tend to exhibit unusually strong cross-modal similarity compared to clean samples, as the contrastive objective encourages the model to tightly bind the trigger with a target concept. CleanCLIP leverages this property by analyzing similarity distributions to detect and suppress suspicious pairs without requiring explicit trigger knowledge.

CleanCLIP proposes a two-stage defense framework. In the first stage, it performs similarity-based filtering to identify candidate poisoned pairs. Samples whose image-text similarity significantly deviates from the expected distribution are flagged as suspicious. In the second stage, CleanCLIP applies targeted data augmentation and reweighting to reduce the influence of these suspicious samples during contrastive training. Rather than outright removing data—which risks harming representation quality—the method softly attenuates their contribution to the loss function.

Experimental results show that CleanCLIP effectively reduces attack success rates for a variety of multimodal backdoor attacks, including patch-based and semantic triggers, while largely preserving clean performance on zero-shot and fine-tuned tasks. The defense is lightweight and integrates easily into standard CLIP training pipelines, making it practical for large-scale deployment.

However, CleanCLIP has several limitations. Its effectiveness depends on the assumption that poisoned samples exhibit detectable similarity anomalies, which may not hold for adaptive attacks designed to mimic natural semantic correlations. Subsequent studies demonstrate that carefully optimized triggers can evade similarity-based filtering. Additionally, CleanCLIP primarily operates at the pair level and does not explicitly address representation-level shortcut learning, which may allow residual backdoor behavior to persist.

Despite these limitations, CleanCLIP represents a foundational contribution to the study of backdoor defenses in contrastive learning. It establishes the feasibility of data-centric defenses in multimodal self-supervised learning and motivates subsequent work, such as semantic counterfactual augmentation, that targets deeper representation-level vulnerabilities.

%CleanerCLIP~\cite{xun2024cleanerclip}
\subsection{CleanerCLIP~\cite{xun2024cleanerclip}}

Recent studies have shown that multimodal contrastive models such as CLIP are particularly vulnerable to backdoor attacks, where poisoned image-text pairs implant malicious behaviors that persist across downstream tasks. Unlike supervised learning, backdoors in contrastive learning often exploit semantic shortcuts in representation space, making them difficult to detect using label-based or prediction-based defenses. CleanerCLIP addresses this challenge by proposing a defense mechanism specifically designed for the semantic structure of contrastive learning.

The core insight of CleanerCLIP is that contrastive backdoors rely on spurious semantic correlations between triggers and target concepts. Instead of filtering poisoned data or injecting random noise, CleanerCLIP aims to systematically break these correlations through fine-grained counterfactual semantic augmentation. The method operates during contrastive pretraining and does not require prior knowledge of the trigger pattern, poisoning ratio, or target label, making it largely attack-agnostic.

CleanerCLIP decomposes multimodal representations into fine-grained semantic attributes and identifies dimensions that exhibit abnormally strong correlations. For potentially poisoned samples, the method generates counterfactual semantic variants by selectively modifying semantic attributes while preserving other contextual information. For example, an image containing a visual trigger may be paired with alternative textual descriptions that remove or alter the target concept, or text prompts may be aligned with visually similar but trigger-free images. These counterfactual pairs explicitly contradict the poisoned semantic association.

The generated counterfactual samples are incorporated into contrastive training via a re-regularized loss function that penalizes reliance on trigger-specific correlations while maintaining alignment for genuine semantic features. This forces the model to learn more causally meaningful representations and suppresses shortcut features associated with backdoors. Importantly, the defense preserves the overall contrastive objective and is compatible with standard CLIP-style training pipelines.

Extensive experiments demonstrate that CleanerCLIP significantly reduces attack success rates across a wide range of backdoor attacks, including patch-based, semantic, and multimodal triggers, often reducing attack success from over 90\% to near-random levels. At the same time, the method incurs negligible degradation in clean performance on zero-shot and linear-probe benchmarks. Compared to prior defenses such as data filtering or robust training, CleanerCLIP shows superior effectiveness, particularly against adaptive attacks that exploit semantic plausibility.

Despite its effectiveness, CleanerCLIP introduces additional computational overhead due to counterfactual generation and relies on the quality of semantic decomposition, which may limit applicability in domains with weak semantic structure. Nevertheless, CleanerCLIP represents one of the first defenses explicitly designed to counter representation-level backdoors in contrastive learning, highlighting the importance of semantic- and representation-aware defenses for secure self-supervised learning.

\subsection{Comparison of CleanCLIP vs. CleanerCLIP}

in this section,we compare the various defense techniques used to counter the backdoor attacks on contrastive learning. Table~\ref{tab:cleanclip_vs_cleanerclip} summarizes the key differences between the two approaches across multiple dimensions relevant to contrastive backdoor defense.

\begin{table*}[t]
\scriptsize
\centering
\caption{Comparison of CleanCLIP and CleanerCLIP}
\label{tab:cleanclip_vs_cleanerclip}
\begin{tabular}{lcc}
\toprule
\textbf{Aspect} & \textbf{CleanCLIP} & \textbf{CleanerCLIP} \\
\midrule
Defense Granularity & Pair-level (image--text similarity) & Representation-level (semantic attributes) \\
Primary Mechanism & Similarity-based filtering and reweighting & Counterfactual semantic augmentation \\
Robustness to Adaptive Attacks & Moderate; can be bypassed by semantic triggers & High; breaks trigger--target correlations \\
Computational Overhead & Low; minimal impact on training & Moderate to High; requires counterfactual generation \\
Effect on Clean Performance & Minimal degradation & Minimal to slight improvement \\
Threat Assumptions & Detectable anomalies in poisoned pairs & Exploits spurious semantic correlations; attack-agnostic \\
Compatibility & Standard CLIP pipelines & Standard CLIP pipelines; extendable to other contrastive models \\
Scalability & High; lightweight for large datasets & Dependent on quality of semantic decomposition; may be slower for massive datasets \\
Industrial/Federated Applicability & Limited; best for centralized datasets & Better; can be adapted to federated or industrial scenarios \\
\bottomrule
\end{tabular}
\end{table*}

CleanCLIP and CleanerCLIP represent two closely related yet fundamentally different defense philosophies for mitigating backdoor attacks in contrastive learning, particularly in multimodal CLIP-style models. While both methods aim to reduce the impact of poisoned image--text pairs without relying on labeled data, they differ substantially in their threat assumptions, defense granularity, and robustness against adaptive attacks.

CleanCLIP adopts a \emph{data-centric, pair-level} defense strategy. Its core assumption is that poisoned image--text pairs exhibit abnormal similarity patterns in the joint embedding space due to the contrastive objective excessively binding triggers to target concepts. By detecting and suppressing such anomalous pairs through similarity-based filtering and reweighting, CleanCLIP effectively reduces attack success rates for a wide range of non-adaptive backdoor attacks. This approach is lightweight, easy to integrate into existing CLIP pipelines, and incurs minimal computational overhead. However, its effectiveness relies on the detectability of poisoned pairs, making it vulnerable to adaptive attacks that carefully craft triggers to mimic natural semantic correlations.

In contrast, CleanerCLIP operates at a \emph{semantic and representation level}, targeting the root cause of contrastive backdoors rather than their surface manifestations. Instead of identifying poisoned samples directly, CleanerCLIP breaks spurious trigger--target correlations by introducing fine-grained counterfactual semantic augmentations during training. This strategy forces the model to rely on causally meaningful semantic features rather than shortcut associations. As a result, CleanerCLIP demonstrates stronger robustness against adaptive attacks and significantly reduces backdoor transferability across downstream tasks. The trade-off is increased computational cost and reliance on effective semantic decomposition, which may limit scalability in extremely large or weakly structured datasets.

From a security perspective, CleanCLIP can be viewed as an effective \emph{first-line defense} against opportunistic or non-adaptive poisoning attacks, while CleanerCLIP provides a more \emph{principled and resilient defense} against sophisticated adversaries. Their complementary strengths suggest that hybrid defenses—combining similarity-based filtering with semantic counterfactual regularization—may offer improved robustness in future contrastive learning systems.